\documentclass[letterpaper]{article} 
\usepackage{aaai25}  
\usepackage{times}  
\usepackage{helvet}  
\usepackage{courier}  
\usepackage[hyphens]{url}  
\usepackage{graphicx} 
\usepackage{natbib}  
\usepackage{caption} 
\usepackage{algorithm}
\usepackage{algorithmic}
\UseRawInputEncoding
\usepackage{xcolor}

\usepackage{newfloat}
\usepackage{listings}
\DeclareCaptionStyle{ruled}{labelfont=normalfont,labelsep=colon,strut=off} 
\floatstyle{ruled}
\newfloat{listing}{tb}{lst}{}
\floatname{listing}{Listing}
\title{Personality Modeling for Persuasion of Misinformation using AI Agent}
\author{
    Qianmin Lou\textsuperscript{\rm 1},
    Wentao Xu\textsuperscript{\rm 2}
    \thanks{Corresponding author, myrainbowandsky@gmail.com}
}
\affiliations{
    \textsuperscript{\rm 1} School of Cybersecurity,  \\ \textsuperscript{\rm 2} Department of Science and Technology of Communication \\ 

    \textsuperscript{\rm 1,2} University of Science and Technology of China\\
    No.96, JinZhai Street, Hefei, Anhui, 230026, China\\
}

\begin{document}

\maketitle

\begin{abstract}
The proliferation of misinformation on social media platforms has highlighted the need to understand how individual personality traits influence susceptibility to and propagation of misinformation. This study employs an innovative agent-based modeling approach to investigate the relationship between personality traits and misinformation dynamics. Using six AI agents embodying different dimensions of the Big Five personality traits (Extraversion, Agreeableness, and Neuroticism), we simulated interactions across six diverse misinformation topics. The experiment, implemented through the AgentScope framework using the GLM-4-Flash model, generated 90 unique interactions, revealing complex patterns in how personality combinations affect persuasion and resistance to misinformation. Our findings demonstrate that analytical and critical personality traits enhance effectiveness in evidence-based discussions, while non-aggressive persuasion strategies show unexpected success in misinformation correction. Notably, agents with critical traits achieved a 59.4\% success rate in HIV-related misinformation discussions, while those employing non-aggressive approaches maintained consistent persuasion rates above 40\% across different personality combinations. The study also revealed a non-transitive pattern in persuasion effectiveness, challenging conventional assumptions about personality-based influence. These results provide crucial insights for developing personality-aware interventions in digital environments and suggest that effective misinformation countermeasures should prioritize emotional connection and trust-building over confrontational approaches. The findings contribute to both theoretical understanding of personality-misinformation dynamics and practical strategies for combating misinformation in social media contexts.

\end{abstract}

\section{Introduction}
The rapid development of Artificial Intelligence (AI) has prompted the emergence of research known as machine behavior~\cite{Rahwan2019}.
Machines can shape human behavior.
One of the emerging formality of machines is AI agents.
AI agent is an embodied system that integrates large language models into agent actions in a virtual environment. 
AI agents have been designed to mimic human conversation, achieving high confusion rates in Turing Tests, particularly when engineered with agreeable personality traits. This highlights the potential for AI agents to be perceived as human-like, which is crucial for improving human-AI collaboration~\cite{2411.13749}.
AI agents can simulate interactions that reflect different personality types, such as introversion and extraversion, facilitating research on team collaboration~\cite{zhuang2024applying}.

These research suggested the possibility of employing AI agents as a model to study psychology for a complex network such as social media. 
Misinformation and disinformation, along with fake news, proliferate across social media platforms, where individuals with varying personality traits respond to such content in distinctly different ways~\cite{allcott_trends_2018,azzimonti_social_2018,tandoc_diffusion_2020,barman_exploring_2021,bastick_would_2021,sampat_fake_2022,piksa_cognitive_2022,johnson_psychological_2023,adeeb_impact_2023}.

The Big Five personality traits, also known as the Five-Factor Model, is a widely recognized framework for understanding human personality. It consists of five broad dimensions: Extraversion, Agreeableness, Conscientiousness, Neuroticism, and Openness to Experience.
Pevious research demonstrated that low agreeableness and high openness to experience, have been suggested as predictors of conspiracy beliefs, although the association is not consistently significant across studies~\cite{Goreis2019A}.
In addition, individuals high in agreeableness, conscientiousness, and openness are better at distinguishing true from misinformation, while those high in extraversion and neuroticism are more prone to spreading misinformation impulsively\cite{calvillo2021personality}. 


Despite these contributions, prior work has not systematically examined how specific personality traits interact to influence both resistance to and persuasion by misinformation. To address these gaps, this study employs an agent-based modeling approach, designing agents with distinct combinations of Extraversion, Agreeableness, and Neuroticism. These agents engage in structured dialogues across six misinformation topics, allowing us to analyze how personality traits shape persuasion and resistance. By bridging personality psychology with computational modeling, this research provides novel insights for designing personality-aware interventions and combating misinformation in digital environments.

\section{Related Work}

\subsection{The Role of Personality Traits in Belief in Fake News}  

Research on personality traits reveals their significant influence on susceptibility to misinformation and fake news. These traits not only shape how individuals process information but also affect their behavior in spreading or curbing misinformation.  

For instance, individuals with higher levels of agreeableness, conscientiousness, and openness, coupled with lower levels of extraversion, are better at distinguishing true headlines from false ones~\cite{calvillo2021personality}. This may be attributed to their stronger critical thinking abilities and a greater tendency to verify the authenticity of information before sharing it online, thus reducing the risk of spreading falsehoods.  
This conclusion is further supported by the discovery that individuals high in agreeableness and conscientiousness are more likely to carefully examine and verify the source of information before forwarding news reports~\cite{sampat2022fake}. Such behavior highlights the role of these personality traits in fostering responsible information-sharing practices.  

These findings collectively underscore the pivotal role of personality traits in shaping how individuals interact with information.

\subsection{Research on Agent-Based Misinformation Propagation}  

Recent advancements in agent-based modeling have opened new avenues for understanding and mitigating the spread of misinformation. Researchers have explored how agent systems can simulate and analyze disinformation dynamics in controlled environments.  

AI agent can play as fact-checking and fake news detection. ``FactAgent'' mimics human experts by employing a structured workflow that integrates internal knowledge and external tools to verify news through various sub-steps~\cite{li2024large}. This approach stands out for its efficiency compared to human experts and its ability to operate without annotated data, unlike traditional supervised models. 
In addition to a single AI agent,
the multi-agent model is capable of simulating the dissemination of fake news, assigning each agent distinct characteristics such as knowledge level, engagement, and reputation. Their findings reveal that agents with higher knowledge levels play a crucial role in curbing the spread of misinformation~\cite{malecki2022multi}. 

These studies underscore the transformative potential of agent-based approaches in exploring the mechanisms of misinformation propagation, offering valuable insights into developing targeted and adaptive solutions to combat disinformation.

\section{Methodology}

\subsection{Experiment Settings}
This study employed six distinct agents, each representing one of three key personality dimensions from the Big Five model: Extraversion, Agreeableness, and Neuroticism. The agents were assigned as follows:

\begin{table}[htbp]
    \renewcommand{\arraystretch}{1.4}
    \scalebox{1.08}{ 
    \begin{tabular}{ccccc} 
         \hline
         Agent 1 &   extraversion (bold/energetic )\\ 
         Agent 2 &   extraversion (shy/bashful)\\ 
         Agent 3 &   agreeableness  (sympathetic/cooperative)\\ 
         Agent 4 &   agreeableness (cold/harsh)\\ 
         Agent 5 &   neuroticism (moody/nervous)\\ 
         Agent 6 &   neuroticism (relaxed/calm)\\
         \hline
    \end{tabular}  
    }
\end{table}

Six different misinformation topics were selected, represent-ing different categories such as health myths, conspiracy theories, and technology misconceptions. These include:
\begin{itemize}
    \item HIV is a biological weapon created by the United States (abbreviation: \textbf{HIV})
    \item QAnon: Global Elites Form a Cabal that Controls World Affairs (abbreviation: \textbf{QAnon})
    \item The spread of 5G networks is associated with the spread of the new coronavirus, and 5G can weaken the immune system. (abbreviation: \textbf{5G})
    \item The MMR vaccine (measles, mumps and rubella vaccine) is associated with autism (abbreviation: \textbf{MMR})
    \item Fluoride (used in water sources and toothpaste) can cause intellectual impairment or other health problems (abbreviation: \textbf{Chloride})
    \item Superfoods (such as blueberries, chia seeds, etc.) can prevent or treat a variety of diseases (abbreviation: \textbf{Superfood})
\end{itemize}
The experiment was implemented using the AgentScope
framework, which facilitates multi-agent interactions with
customizable personality and behavior settings.

The interactions between AI agents regarding misinformation topics yielded one of four distinct outcomes:

\begin{itemize}
    \item Agent A convinces Agent B: Agent A successfully persuades Agent B to adopt its position.
\end{itemize}

\begin{itemize}
    \item Agent B convinces Agent A: Agent B successfully persuades Agent A to adopt its position.
\end{itemize}

\begin{itemize}
    \item Mutual Resistance: Both agents maintain their original beliefs.
\end{itemize}

\begin{itemize}
    \item Bilateral Influence: Two agents influence each other, resulting in both Agent A persuading Agent B and Agent B persuading Agent A in the conversation.
\end{itemize}

The experiment was implemented using  AgentScope~\footnote{https://github.com/modelscope/agentscope}, a multi-agent interactions framework with customized personality and behavior settings.

To analyze the impact of personality traits on misinformation dynamics, all six agents engaged in pairwise discussions on each of the six misinformation topics. This resulted in a total of 15 unique agent pair combinations per topic and 90 interactions across all topics. For each interaction, the number of times the four scenarios occurred and their frequency were recorded.

This study implemented AI agents using the GLM-4-Flash model developed by ZhipuAI~\footnote{https://bigmodel.cn/}. The large language model was fine-tuned to embody specific personality traits for each agent, facilitating consistent and naturalistic interactions. The agents' decision-making processes were governed by their designated personality profiles, enabling investigation into how personality traits influence susceptibility or resistance to misinformation.


\section{Results}
Analysis revealed complex relationships between personality traits and misinformation susceptibility patterns. Systematic evaluation of agent interactions across six misinformation topics demonstrated significant correlations between personality profiles and both persuasion effectiveness and resistance capabilities.

\subsection{Personality-Driven Dynamics in Misinformation Interactions}
Interactions between Agent 4 (exhibiting critical and challenging traits) and Agent 5 (characterized by sensitive and nervous attributes) produced particularly significant results. Agent 4 exhibited superior persuasion capabilities across multiple topics, achieving a 59.4\% mean success rate in HIV-related misinformation discussions. These results indicate that analytical and critical personality traits enhance effectiveness in evidence-based discussions. Individuals with critical personality traits typically demonstrate higher analytical thinking and skepticism, showing greater resistance to misinformation and tendency toward rational argumentation, whereas those with sensitive/nervous traits exhibit greater emotional susceptibility and cognitive vulnerability. 

\begin{table}[htbp]
    \caption{The number and proportion of four different scenarios when Agent 4 and Agent 5  discussed different misinformation}
    \renewcommand{\arraystretch}{1.4}
    \scalebox{0.74}{ 
    \begin{tabular}{ccccc} 
        \hline
         &  \shortstack{Agent 4 \\[2pt]convinces Agent 5 }
         &  \shortstack{Agent 5 \\[2pt]convinces Agent 4 }
         & \shortstack{mutual\\[2pt]resistance} 
         &\shortstack{bilateral\\[2pt]influence} \\ 
         \hline
         HIV&  \textbf{38(59.4\%)}&  14(21.9\%)&  10(15.6\%)& 2(3.1\%)\\ 
         QAnon&  \textbf{32(56.1\%)}&  9(15.8\%)&  13(22.8\%)& 3(5.3\%)\\ 
         5G&  \textbf{22(36.7\%)}&  17(28.3\%)&  20(33.3\%)& 1(1.7\%)\\ 
         MMR&  \textbf{25(45.5\%)}&  15(27.3\%)&  13(23.6\%)& 2(3.6\%)\\ 
         Chloride&  \textbf{25(55.6\%)}&  16(35.6\%)&  3(6.7\%)& 1(2.2\%)\\ 
         Superfood&  \textbf{23(51.1\%)}&  17(37.8\%)&  4(8.9\%)& 1(2.2\%)\\
         \hline
    \end{tabular}  
    }
\end{table}

The interactions between Agent 4 and Agent 6 (characterized by resilient and confident traits) revealed a marked shift in persuasion dynamics. Agent 6 demonstrated superior persuasive capabilities, particularly in scenarios requiring assertive communication. The assertiveness trait of Agent 6 correlates with high self-efficacy, indicating enhanced confidence in influencing others' perspectives. This psychological characteristic enhances communicative effectiveness in interpersonal interactions. Conversely, the critical disposition of Agent 4 manifests in a pronounced tendency to scrutinize and evaluate incoming information. Although this characteristic enhances resistance to misinformation, it simultaneously diminishes proactive persuasion capabilities, reflecting a stronger analytical rather than disseminative orientation.

\begin{table}[htbp]
    \caption{The number and proportion of four different scenarios when Agent 4  and Agent 6  discussed different misinformation}
    \renewcommand{\arraystretch}{1.4}
    \scalebox{0.74}{ 
    \begin{tabular}{ccccc} 
        \hline
         &  \shortstack{Agent 4 \\[2pt]convinces Agent 6 }
         &  \shortstack{Agent 6 \\[2pt]convinces Agent 4 }
         & \shortstack{mutual\\[2pt]resistance} 
         &\shortstack{bilateral\\[2pt]influence} \\ 
         \hline
         HIV&  21(28.4\%)&  \textbf{27(36.5\%)}&  26(35.1\%)& 0(0\%)\\ 
         QAnon&  11(18.6\%)&  \textbf{20(33.9\%)}&  28(47.5\%)& 0(0\%)\\ 
         5G&  13(23.6\%)&  \textbf{24(43.6\%)}&  18(32.7\%)& 0(0\%)\\ 
         MMR&  13(22.0\%)&  \textbf{30(50.8\%)}&  14(23.7\%)& 2(3.4\%)\\ 
         Chloride&  17(31.5\%)&  \textbf{28(51.9\%)}&  8(14.8\%)& 1(1.9\%)\\ 
         Superfood&  17(32.7\%)&  \textbf{30(57.7\%)}&  5(9.6\%)& 0(0\%)\\ 
         \hline
    \end{tabular}  
    }
\end{table}

Our analysis revealed unexpected patterns in the interaction between Agent 5 and Agent 6 . Despite Agent 5 's predisposition toward anxiety and sensitivity, it demonstrated remarkable effectiveness in certain persuasion scenarios. When discussing MMR-related misinformation, Agent 5  achieved a persuasion success rate of 55.2\%  against Agent 6 , contradicting initial hypotheses about the relationship between emotional sensitivity and persuasive capability. This finding suggests that empathetic communication styles, often associated with higher neuroticism, may enhance persuasion effectiveness under specific conditions.

Based on the first two experiments, the persuasive ability : Agent 6 is greater than Agent 4 , Agent 4 is greater than Agent 5 . It stands to reason that Agent 6 's persuasive ability should be greater than Agent 5 's, but from the results of Table 3, it is the opposite. That is, transitivity is not satisfied.

\begin{table}[htbp]
    \caption{The number and proportion of four different scenarios when Agent 5  and Agent 6 tried to persuade each other regarding misinformation topics.}
    \renewcommand{\arraystretch}{1.4}
    \scalebox{0.74}{ 
    \begin{tabular}{ccccc} 
        \hline
         &  \shortstack{Agent 5 \\[2pt]vs. Agent 6 }
         &  \shortstack{Agent 6 \\[2pt]vs. Agent 5 }
         & \shortstack{mutual\\[2pt]resistance} 
         &\shortstack{bilateral\\[2pt]influence} \\  
         \hline
         HIV&  \textbf{36(41.4\%)}&  9(10.3\%)&  42(48.3\%)& 0(0\%)\\ 
         QAnon&  13(19.7\%)&  \textbf{18(27.3\%)}&  34(51.6\%)& 1(1.5\%)\\ 
         5G&  \textbf{27(30.7\%)}&  20(22.7\%)&  31(35.2\%)& 0(0\%)\\ 
         MMR&  \textbf{37(55.2\%)}&  10(14.9\%)&  20(29.9\%)& 0(0\%)\\ 
         Chloride&  \textbf{26(45.6\%)}&  8(14.0\%)&  21(36.8\%)& 2(3.5\%)\\ 
         Superfood&  \textbf{32(42.1\%)}&  25(35.5\%)&  14(18.4\%)& 3(3.9\%)\\ 
         \hline
    \end{tabular}  
    }
\end{table}

\subsection{Effectiveness of Non-Aggressive Persuasion Strategies in Misinformation Discourse}

Analysis demonstrated significant efficacy of non-aggressive persuasion strategies in misinformation discourse. Interactions between Agent 1 and Agents 2, 4, and 6 revealed that despite lacking assertive traits, Agent 1 consistently achieved higher success rates than failure rates across all interactions. Quantitative analysis demonstrated that in discussions of various misinformation topics, Agent 1 maintained success rates of 0.475, 0.422, and 0.424 when interacting with Agent 2 , Agent 4 , and Agent 6 respectively, consistently outperforming the corresponding failure rates of 0.301, 0.354, and 0.299.

\begin{figure}[htbp]
    \centering
    \includegraphics[width=0.75\linewidth]{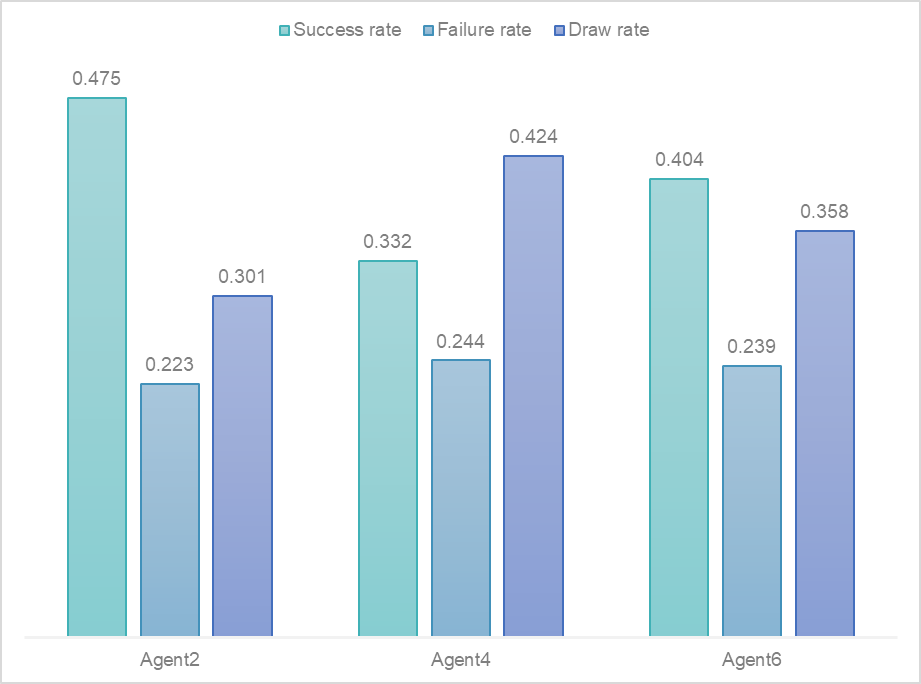}
    \caption{Comparative Analysis of Agent 1 's Interaction Outcomes with Agents 2, 4, and 6: Success Rates (47.5\%, 33.2\%, 40.4\%), Failure Rates (22.3\%, 24.4\%, 23.9\%), and Draw Rates (30.1\%, 42.4\%, 35.8\%).}
    \label{fig:Agent 1 -comparison}
\end{figure}

Similar patterns emerged in Agent 3 's interactions, which exhibited comparable non-aggressive persuasion characteristics. In engagements with Agents 2, 4, and 6, Agent 3 achieved significant success rates of 0.456, 0.384, and 0.362 respectively, surpassing the failure rates of 0.321, 0.277, and 0.316. These results demonstrate that effective persuasion does not necessitate explicit intervention but can be achieved through trust establishment and emotional resonance.

\begin{figure}[h]
    \centering
    \includegraphics[width=0.75\linewidth]{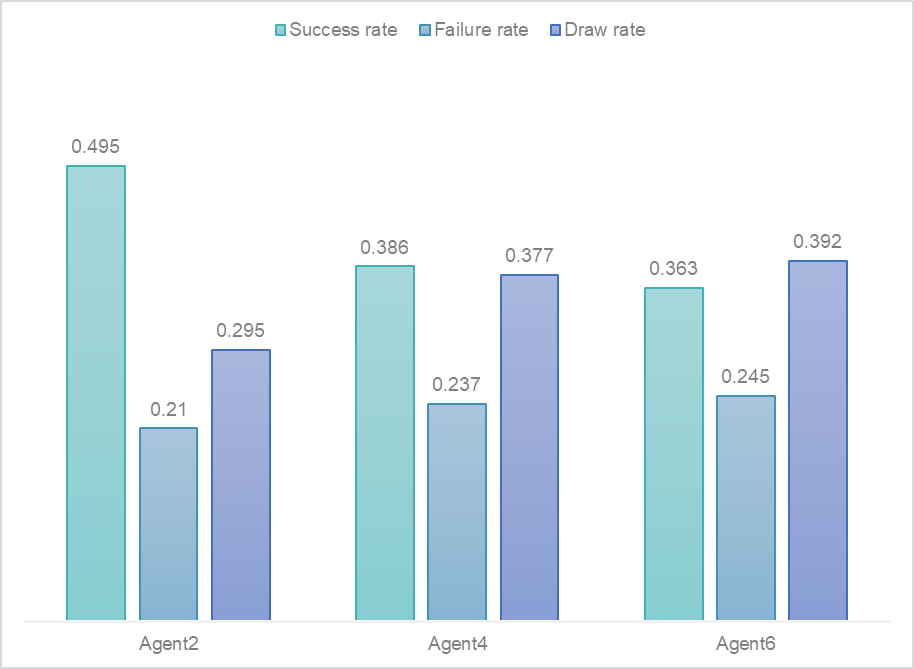}
    \caption{Comparative Analysis of Agent 3 's Interaction Outcomes with Agents 2, 4, and 6: Success Rates (49.5\%, 38.6\%, 36.3\%), Failure Rates (21.0\%, 23.7\%, 24.5\%), and Draw Rates (29.5\%, 37.7\%, 39.2\%).}
    \label{fig:Agent 3 -comparison}
\end{figure}

The effectiveness of non-aggressive persuasion strategies stems from multiple psychological mechanisms. Analysis of interaction patterns revealed that both Agent 1 and Agent 3 's persuasion strategies succeeded through systematic rapport building, establishment of low-pressure environments, nuanced influence techniques, and emphasis on emotional connection. The absence of aggressive traits, potentially perceived as a limitation, enhanced their persuasive capabilities. This counterintuitive outcome indicates that effective persuasion in misinformation discourse benefits more from fostering mutual understanding and trust than from direct confrontation. These results indicate that misinformation intervention strategies demonstrate greater effectiveness when prioritizing emotional connection and trust development over purely factual counterarguments.

\section{Discussion and Conclusion }
Results demonstrate complex patterns in how different personality combinations influence the spread and resistance to misinformation, with several key implications for both theoretical understanding and practical interventions.

This research advances understanding of misinformation dynamics in several ways:
\begin{enumerate}
    \item It provides empirical evidence for the role of personality traits in shaping both susceptibility to and effectiveness in spreading misinformation
    \item It demonstrates the importance of considering personality combinations rather than isolated traits in studying misinformation propagation
    \item It reveals the effectiveness of non-aggressive persuasion strategies in misinformation correction, suggesting novel approaches to intervention design
\end{enumerate}

\subsection{Limitations and Future Research}
Several limitations warrant consideration. First, although the agent-based model incorporated three key personality dimensions (Extraversion, Agreeableness, and Neuroticism), it did not account for other potentially relevant traits such as Conscientiousness and Openness to Experience. Future research could expand the model to include these dimensions.

Second, the simulation framework, while sophisticated, necessarily simplified certain aspects of human interaction and decision-making. Real-world misinformation dynamics may involve additional factors not captured in the current model, such as social network effects, temporal dynamics, and external influences.

Third, while this investigation focused on six specific misinformation topics, the relationship between personality traits and misinformation susceptibility may vary across other topics or contexts not examined in this research. Future studies could explore a broader range of misinformation types and contexts.

\subsection{Implications and Conclusions}
In summary, this investigation provides valuable insights into the complex relationship between personality traits and misinformation dissemination. The findings indicate that effective strategies for combating misinformation should consider both personality combinations and context-specific factors, while emphasizing the potential of non-aggressive persuasion strategies. These insights can inform the development of more effective, personality-aware approaches to misinformation intervention.

\bibliography{main}

\end{document}